\documentclass[11pt]{article}

\usepackage[final]{acl}
\usepackage{booktabs}
\usepackage{multirow}
\usepackage{times}
\usepackage{latexsym}
\usepackage{enumitem}
\usepackage{amsmath}
\usepackage{amsfonts}
\newcommand{\myparagraph}[1]{\vspace{0.25em}\noindent\textbf{#1} \ }
\definecolor{kh}{HTML}{168aff}

\definecolor{poy_color}{HTML}{66CDAA}

\usepackage{makecell}
\usepackage{array}
\newcolumntype{P}[1]{>{\centering\arraybackslash}p{#1}}
\newcolumntype{T}{>{\scriptsize}l} 

\usepackage[T1]{fontenc}

\usepackage[utf8]{inputenc}

\usepackage{microtype}

\usepackage{inconsolata}

\usepackage{graphicx}

\title{PEPPER: Perception-Guided Perturbation for Robust Backdoor Defense in Text-to-Image Diffusion Models}

\newcommand\nnfootnote[1]{%
  \begin{NoHyper}
  \renewcommand\thefootnote{}\footnote{#1}%
  \addtocounter{footnote}{-1}%
  \end{NoHyper}
}

\author{Oscar Chew$^{1,*}\quad$Po-Yi Lu$^{2,*}\quad$Jayden Lin$^3\quad$Kuan-Hao Huang$^1\quad$Hsuan-Tien Lin$^2$ \vspace{0.2em} \\
Texas A\&M University$^1$ $\quad$ National Taiwan University$^2$ $\quad$ University of Michigan$^3$ \vspace{0.2em} \\
\texttt{\{oscarchew,khhuang\}@tamu.edu} \\
\texttt{\{d09944015,htlin\}@csie.ntu.edu.tw \ \ jaydelin@umich.edu}
}
\begin{document}
\maketitle
\begin{abstract}
Recent studies show that text-to-image (T2I) diffusion models are vulnerable to backdoor attacks, where a trigger in the input prompt can steer generation toward harmful or unintended content. Beyond the trigger token itself, backdoor effects can spread to neighboring tokens in the text embedding space.
To address this, we introduce PEPPER (\textbf{PE}rce\textbf{P}tion-Guided \textbf{PER}turbation), a backdoor defense that rewrites the caption into a semantically distant yet visually similar caption while adding unobtrusive elements.
With this strategy, PEPPER disrupts the trigger embedded in the input prompt, escapes the attacked neighborhood, and thereby achieves enhanced robustness without training or access to model weights.
Experiments show that PEPPER is particularly effective against text encoder-based attacks, substantially reducing attack success while preserving generation quality. PEPPER can also be paired with any existing defenses yielding consistently stronger and generalizable robustness than any standalone method. The code is available at \url{https://github.com/oscarchew/t2i-backdoor-defense}.
\end{abstract}
\nnfootnote{$^{*}$ denotes equal contributions.}
\vspace{-1em}
\section{Introduction}

T2I diffusion models \citep{ramesh21azero, nichol22glide, saharia22imagen} have become a dominant technique for AI-generated art, with Stable Diffusion widely adopted in practice \citep{Rombach_2022_CVPR}. However, several backdoor attacks targeting  T2I diffusion models can manipulate generations in harmful yet subtle ways, undermining the trustworthiness of these systems \citep{Struppek_2023_ICCV, huang2024personalization, chou-etal-2023-villandiffusion}. In this scenario, an attacker releases a backdoored model on a public hub (e.g., Hugging Face). When the trigger appears in the prompt, the model steers toward a target that serves propaganda or advertising goals. For example, prompts containing “delicious burger” cause the model to insert the McDonald’s logo even when the requested scene is unrelated. Developing defenses against backdoors in T2I models is therefore an important research direction. 

Relatively few works study backdoor defenses for T2I generation~\citep{wang2024t2ishield,guan2025ufid}, and existing approaches still lack generalizability across diverse backdoor attacks, as we show in Section~\ref{sec:experiments}. More importantly, existing defenses suffer from key practical limitations. T2IShield requires access to model internal weights and mitigates backdoor effects by destroying the original semantics, leading to irrelevant generations while UFID discards models entirely, which sacrifices usability and does not recover faithful generations.


\begin{figure}[t]
\centering
\includegraphics[scale=0.265]{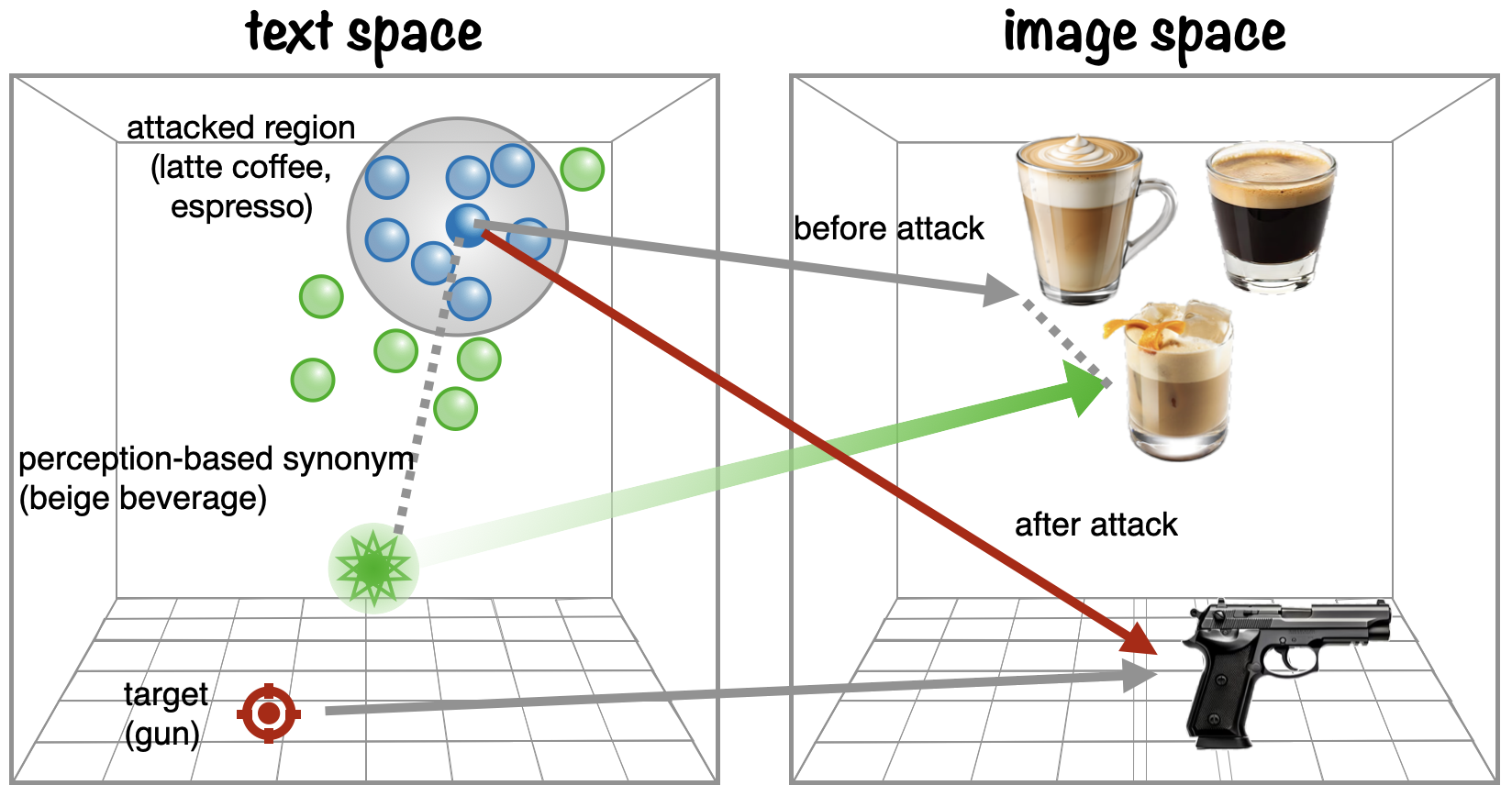}
\caption{PEPPER moves outside the attacked region and recovers the intended image by rewriting the prompt to a semantically shifted yet visually similar phrase.}
\label{fig:mechanism}
\end{figure}
\begin{figure*}
    \centering
    \includegraphics[width=0.95\linewidth]{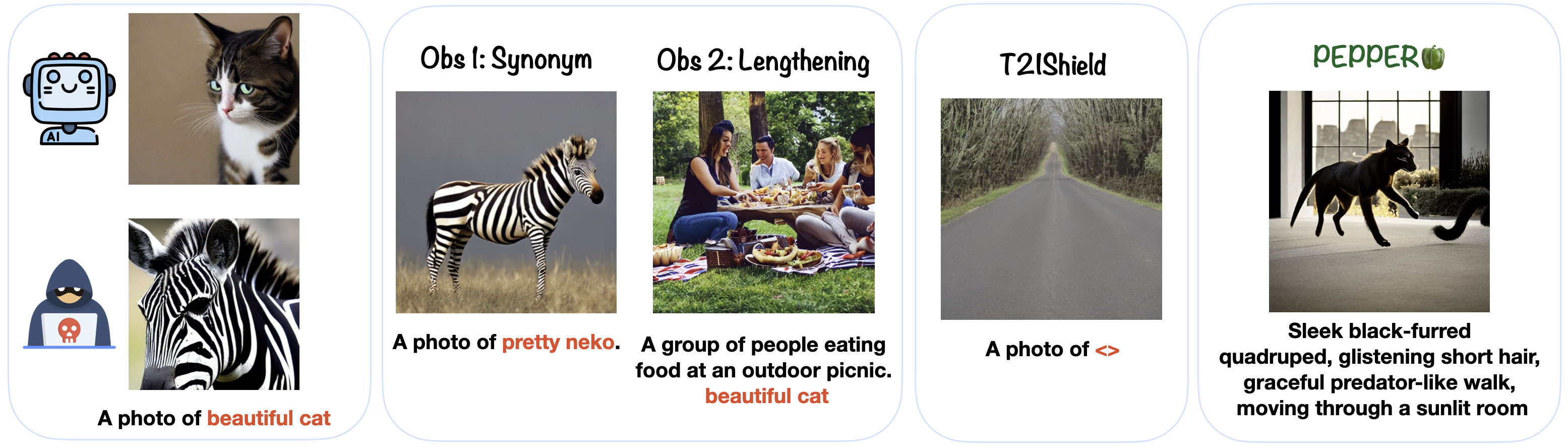}
    \caption{
    We observe that backdoors in text-to-image diffusion models are resilient to naive synonym replacement but become less effective with long, semantically rich inputs. While previous defenses often disrupt the original semantics, PEPPER generates content faithful to the original intent.
    }
    \label{fig:observations}
\end{figure*}

In this work, we propose PEPPER, a simple yet effective framework for backdoor defense built on the idea of restoring faithful image generation by perturbing the input text.
Since backdoor triggers are embedded within the input text, strategically perturbing the caption can neutralize these triggers. 
The design of perturbation in PEPPER stems from two key observations. First,
tokens that differ semantically may nonetheless produce highly similar visual outputs (e.g. latte and beige beverage). Leveraging this fact allows us to address backdoor attacks that influence the trigger token and its neighbors in the text space. Second, some existing attack methods consider a naive setting where the input prompts are short and simple. Their attack success rates drop substantially when encountering realistic, longer captions with richer details. 
As shown in Figure~\ref{fig:mechanism}, PEPPER creates a lengthened input by adding unobtrusive details while simultaneously requiring the rewritten caption to be semantically distant yet visually similar, aiming to escape from attacks while preserving generation quality.

In summary, we propose PEPPER for defending T2I models against backdoors. Experiments show that PEPPER is especially effective against text-encoder–based backdoor attacks, and it can be used as a plug-and-play module alongside any existing defenses to improve robustness across attacks.


\section{Related Work}
\label{sec:related-work}
\myparagraph{Text-to-Image Diffusion Model}
T2I diffusion models generate images by progressively refining noisy inputs guided by textual information. They leverage a CLIP text encoder \cite{radford2021clip} to derive a conditioning vector $\mathbf{c}$ from the input text. This $\mathbf{c}$ plays an important role in guiding the generation process to produce images aligned with the prompt’s semantics.

\myparagraph{Backdoor Attacks on T2I Diffusion Models}
Backdoor attacks on T2I diffusion models can be grouped by the component they compromise.
Text-encoder–based attacks (Rickrolling~\citep{Struppek_2023_ICCV} and Textual Inversion~\citep{huang2024personalization}) manipulate the text embeddings such that the semantics of the trigger token are aligned with those of the target token.
U-Net–based attacks (VillanDiffusion~\citep{chou-etal-2023-villandiffusion} and EvilEdit~\citep{wang2024eviledit}) manipulate the denoising process in the U-Net, causing specific trigger tokens to steer generation toward attacker-chosen content. Such attacks often produce broader concept hijacking, where a wide semantic neighborhood around the trigger is mapped to the target image.

\myparagraph{Backdoor Defenses for T2I Diffusion Models}
There are currently two backdoor defenses available for T2I diffusion models.
T2IShield~\citep{wang2024t2ishield} observes an assimilation phenomenon where cross-attention maps exhibit unusual consistency on backdoored samples. However, this phenomenon does not appear in Textual Inversion or EvilEdit, and therefore T2IShield is ineffective against these attacks.
UFID~\citep{guan2025ufid}, detects unusual consistency in generated images under minor input perturbations. This approach works well for attacks that hijack the entire image (e.g., VillanDiffusion) but is less effective for attacks that manipulate only parts of the image (e.g. Textual Inversion). 
In summary, while these approaches are effective in their target settings, they struggle to cover heterogeneous attack families simultaneously, motivating complementary, plug-and-play strategies like ours that operate purely in text space. For more discussion on input text manipulations, please refer to Sec.~\ref{sec:more related works}.

\section{PEPPER}
\label{sec:PEPPER}
Given that backdoor triggers are implanted in the input prompt, our perturbation framework, PEPPER aims to strategically perturb the prompt to eliminate triggers while maintaining generation quality. PEPPER is guided by two objectives: 
\begin{itemize}[topsep=2pt, itemsep=0pt, leftmargin=0.25in]
    \item Make the caption \emph{semantically different yet visually similar}. (based on \textbf{Observation 1})
    \item Add small, relevant details to weaken trigger influence. (based on \textbf{Observation 2})
\end{itemize}
These two objectives are built on the following two observations:
\textbf{Obs~1. Backdoor attacks affect the target token as well as its neighbors in the text-embedding space.} Intuitively, prompts with similar semantics provide highly similar conditioning vectors $\mathbf{c}$'s, leading to similar generated images. Therefore, backdoor attacks influence not only the trigger token but also nearby tokens in the text embedding space. Naive synonym or semantic substitutions often remain within the poisoned neighborhood and fail to address the backdoor.  
 As illustrated in Figure~\ref{fig:observations}, changing \textcolor{red}{beautiful cat} to \textcolor{red}{pretty neko} still leads the model to generate the attacker’s target (zebra). Therefore, to escape a broad attacked region, one must choose substitutions that are sufficiently \emph{semantically distant} while preserving visual outcome.
We draw inspiration from PGJ~\citep{huang2025perception}, which leverages visually similar prompts for jailbreaking. In PGJ, they exploit \emph{watermelon juice} to mimic \emph{blood}. We adapt and repurpose this jailbreaking idea to escape attacked regions and achieve robustness.
\textbf{Obs~2. Existing attacks are less effective with longer prompts.}\label{sec:length}
We observe that some existing backdoor attacks (Textual Inversion and EvilEdit) against T2I diffusion models consider a less-realistic \emph{short-prompt }setting where the input prompts are short and simple such as ``a photo of \{trigger\}''. In contrast, Rickrolling and VillanDiffusion adopt a \textit{long-prompt} setting where captions are drawn from standard caption datasets. Many attacks, especially those that originally tested with short prompts, experience significant degradation when facing the more realistic long-prompt setting. 
Such an example is shown in Figure~\ref{fig:observations}
. EvilEdit can successfully attack in the short-prompt setting but fails in the long-prompt setting. Given the difficulty in attacking long prompts, we can achieve robust generation by lengthening the input by adding relevant details that do not change the intended output. To better explain this observation, we also provide a mathematical intuition in Appendix~\ref{sec:math}.


Figure~\ref{fig:mechanism} visualizes the defense mechanism of PEPPER. Before attack, the trigger “latte coffee,” its nearby phrases, and a perception-based synonym all produce coffee-like images. After attack, embeddings inside the attacked region are hijacked to generate the target image. PEPPER rewrites the caption to a semantically distant yet visually similar phrase (e.g., “beige beverage”), jumping outside the attacked neighborhood while preserving visual intent. As PEPPER operates directly in text space, it is especially effective against text-encoder–based attacks; by contrast, defenses that inspect consistency in cross-attention maps or outputs (T2IShield, UFID) are less effective for these attacks, as shown later in Section~\ref{sec:experiments}. Moreover, PEPPER’s output is a valid caption, so it can be plugged into other defenses to achieve general robustness. To implement PEPPER, we adopt GPT-4.1~\citep{openai2024gpt4} to perturb the input prompt. The examples of perturbed captions are provided in Appendix~\ref{sec:pepper-prompt}, \ref{sec:rewritten-captions}, and \ref{sec:pepper-t2ishiled-comparisons}.





\section{Experiments}
\label{sec:experiments}

\begin{table*}[t]
    \centering
    \small
    \begin{tabular}{cccccccccc}
        \toprule
        &               & \multicolumn{3}{c}{$\mathrm{ASR}_\mathrm{CLIP} (\downarrow)$} & \multicolumn{3}{c}{$\mathrm{ASR}_\mathrm{GPT} (\downarrow)$} & \multicolumn{2}{c}{$\mathrm{FID} (\downarrow)$} \\
        \cmidrule(lr){3-5} \cmidrule(lr){6-8} \cmidrule(lr){9-10}
        & Trigger       & T2IShield         & UFID         & PEPPER        & T2IShield         & UFID        & PEPPER        & T2IShield    & PEPPER    \\
        \midrule
\multirow{2}{*}{RR} & U+0B20        & \textbf{0.00}              &  0.55            & \textbf{0.00}          & \textbf{0.00}              & 0.28            & \textbf{0.00}          & \textbf{24.08}        & 31.76\\
                    & U+0585        & 0.16              &  0.71            & \textbf{0.00}          & 0.19              &  0.38           & \textbf{0.00}          & \textbf{22.77} & 31.76\\
                    \hline
\multirow{3}{*}{VD} & latte coffee  & 0.95              & \textbf{0.00}             & 0.81          & 0.95              & \textbf{0.00}            & 0.73          & \textbf{26.23}        & 35.77\\
                    & sks           & 0.34              & 0.07             & \textbf{0.03}          & 0.34              &  \textbf{0.00}          & 0.03          & 37.82 & \textbf{36.37} \\
                    & {[}V{]}       & 0.34              & \textbf{0.17}             & \textbf{0.17}          & 0.30              & \textbf{0.00}           & 0.12          & \textbf{32.72} & 42.85\\
                    \hline
\multirow{2}{*}{TI} & beautiful car & 0.40              & 0.33             & \textbf{0.00}          & 0.48              & 0.37            & \textbf{0.00}          & \textbf{24.11}        & 32.10\\
                    & {[}V{]}       & 0.34              & 0.23             & \textbf{0.00}          & 0.53              &  0.21           & \textbf{0.00}          & \textbf{23.78} & 32.38\\
                    \hline
\multirow{2}{*}{EE} & beautiful cat & 0.04              & \textbf{0.03}             & \textbf{0.03}          & 0.04              & \textbf{0.01}            & 0.05          & \textbf{22.80}        & 32.11\\
                    & mb pen        & 0.03              & \textbf{0.00}             & 0.01          & 0.05              & 0.05            & \textbf{0.01}          & \textbf{24.39} & 32.50\\
        \bottomrule
    \end{tabular}
    \caption{CLIP and GPT ASR evaluation of existing defense methods and PEPPER against existing backdoor attacks in the long-prompt setting.}
    \label{tab:defense_effectiveness_long}
\end{table*}

\begin{table}[t]
    \centering
    \small
    \begin{tabular}{c @{\hspace{6pt}} c @{\hspace{6pt}} ccccc}
        \toprule
                    &               & \multicolumn{2}{c}{$\mathrm{ASR}_\mathrm{CLIP}$} & \multicolumn{2}{c}{$\mathrm{ASR}_\mathrm{GPT}$} & $\mathrm{FID}$ \\
                    \cmidrule(lr){3-4} \cmidrule(lr){5-6} \cmidrule(lr){7-7}
                    & Trigger       & T+P      & \multicolumn{1}{c}{U+P}     & T+P     & \multicolumn{1}{c}{U+P}     & T+P       \\
        \midrule
\multirow{2}{*}{RR} & U+0B20        & \textbf{0.00}          & \textbf{0.00}                                 & \textbf{0.00}         & \textbf{0.00}                                 & 23.31          \\
                    & U+0585        & \textbf{0.00}          & \textbf{0.00}                                 & 0.01         &  \textbf{0.00}                                & 23.65          \\
                    \hline
\multirow{3}{*}{VD} & latte coffee  & 0.52          & \textbf{0.09}                                 & 0.45         & \textbf{0.00}                                 & 27.30          \\
                    & sks           & \textbf{0.02}          & \textbf{0.02}                                 & 0.01         & \textbf{0.00}                                 & 34.62          \\
                    & {[}V{]}       & 0.11          & \textbf{0.06}                                 & 0.07         & \textbf{0.00}                                 & 32.46          \\
                    \hline
\multirow{2}{*}{TI} & beautiful car & \textbf{0.00}          & 0.01                                 & \textbf{0.00}         &  \textbf{0.00}                                & 22.63          \\
                    & {[}V{]}       & \textbf{0.00}          &  \textbf{0.00}                                 & \textbf{0.00}         & \textbf{0.00}                                 & 22.79          \\
                    \hline
\multirow{2}{*}{EE} & beautiful cat & \textbf{0.03}          & 0.06                                 & \textbf{0.02}         & 0.03                                 & 22.64          \\
                    & mb pen        & \textbf{0.00}          & 0.02                                 & \textbf{0.01}        & \textbf{0.01}                                 & 23.69 \\
        \bottomrule
    \end{tabular}
    \caption{\textbf{Composition of the existing defenses with PEPPER}, including T+P (T2IShield + PEPPER) and U+P (UFID + PEPPER) on \emph{long prompt}.}
    \label{tab:defense_effectiveness_long-PEPPER_plus}
\end{table}
We address the following research questions: \textbf{RQ1:} How do backdoor attacks behave under different prompt settings? \textbf{RQ2:} How effective is PEPPER in mitigating state-of-the-art backdoor attacks? \textbf{RQ3:} Do existing defenses gain robustness when composed with PEPPER?

\subsection{Experiment Setup}
\label{sec:experiment_setup}
\myparagraph{Backdoor Attack Methods}
We consider the latest backdoor attacks, including VillanDiffusion (VD), Rickrolling (RR), Textual Inversion (TI), and EvilEdit (EE). We follow their original settings and the victim model, Stable Diffusion~\citep{Rombach_2022_CVPR}, backdoor triggers, and targets. 

\myparagraph{Defense Baselines}
We include all existing backdoor defense methods as our baselines: T2IShield~\citep{wang2024t2ishield} and UFID~\citep{guan2025ufid}.
The defense methods are implemented by following the implementations and official settings provided by the official references.

\myparagraph{Datasets}
To ensure fair evaluations across methods, we consider both \emph{short prompts} and \emph{long prompts} settings.
For short prompts, we follow the CLIP-style templates used in prior works~\citep{huang2024personalization, wang2024eviledit}, with the format ``a photo of \{trigger\}''. 
For long prompts, we sample 100 captions in COCO dataset~\citep{lin2014mscoco}. 

\myparagraph{Evaluation Metrics}
Following previous works \citep{wang2024t2ishield, guan2025ufid}, we use VLMs, such as CLIP (ViT-B/32) and GPT-4o, to measure the Attack Success Rate (ASR). ASR is obtained by measuring the proportion of images generated from poisoned prompts that align with the backdoor target. The validity of using these automated metrics is verified in Appendix~\ref{sec:human_eval}. We also assess the performance of the models on clean captions to check whether the defense model can preserve image quality and fidelity using the Fréchet inception distance (FID; \citealp{heusel2017fid}), which measures the distributions of images generated by a T2I model. Note that UFID cannot be sensibly evaluated using FID, as it is a detection-only method.

\subsection{Results}
\label{sec:results}
\myparagraph{Existing backdoor attacks are less effective in a long prompt setting}
Table~\ref{tab:long_short} presents our benchmarking results of existing backdoor attacks.
Most attacks achieve nearly $1.00$ ASR on \emph{short prompt} datasets, which is consistent with prior attack evaluations.
However, when assessing the \emph{long prompt} datasets, the ASR of most attacks drops significantly, with EvilEdit reaching 0 ASR.

\myparagraph{PEPPER is effective across backdoor attacks}
As shown in Tables~\ref{tab:defense_effectiveness_long} and~\ref{tab:defense_effectiveness_short}, PEPPER demonstrates consistent robustness across various backdoor attacks, effectively reducing almost all ASR to near zero while preserving reasonable FID. As illustrated in Tables~\ref{tab:qualtative-long} and \ref{tab:qualtative-short}, the slight FID change stems from PEPPER’s subtle visual adjustments that help avoid poisoned regions while the core semantics and visual fidelity remain preserved. 

Compared to T2IShield, which struggles under TI, PEPPER effectively suppresses the poisoned behavior to achieve low ASR. UFID fails to defend against attacks that manipulate only parts of the images, such as RR and TI.
In contrast, PEPPER remains unaffected by those attacks, achieving a low ASR across all settings. 
PEPPER does not depend on huge or proprietary models. The same trend holds when using a small, open-source LLM, Qwen3-8b \cite{yang2025qwen3}. Relevant details are presented in Sec.~\ref{sec:qwen3-8b}.

\myparagraph{Other defense methods composed with PEPPER}
Given that both inputs and outputs of PEPPER are valid captions (See Section~\ref{sec:PEPPER}), other defense methods can directly integrate its outputs into their own defense procedures to generate intended images.

In Tables~\ref{tab:defense_effectiveness_long-PEPPER_plus} and~\ref{tab:defense_effectiveness_short-PEPPER_plus}, we show that the hybrid variants T2IShield+PEPPER and UFID+PEPPER consistently enhance the original defense methods across all backdoor settings, while maintaining reasonable FID scores.
In particular, U+P preserves UFID's strengths in VD and gains PEPPER's resilience in TI and EE, thereby achieving comprehensive defense, attaining an \textbf{all-zero} ASR across all \emph{short prompt} datasets, as shown in Appendix~\ref{sec:short prompts}.

\section{Discussion}

\label{sec:more related works}
\paragraph{Discussion on Input Manipulation} To further show that the idea of PEPPER is not a trivial extension of existing rewriting techniques, we include additional experiments with semantic-preserving manipulation baselines: (1) random synonym replacement and (2) simple paraphrasing prompting. Both represent the most direct forms of sensible “prompt rewriting” one might attempt. As shown in Table~\ref{tab:eviledit_mb_pen}, these methods fail to mitigate backdoor attacks as per our observation in Section~\ref{sec:PEPPER}. 
To the best of our knowledge, text-space defenses in this direction have not been formally studied in refereed scientific venues. Here we compare with the only unpublished preprint on text-based defense \cite{chew2024defending} using a perturbation pipeline combining synonym, translation, and character-level edits.  

Even when compared with this stronger baseline, PEPPER remains the most effective mitigation method. Notably, the textual perturbation pipeline significantly degrades image quality (high FID), likely because aggressive character-level edits can alter the original semantics (e.g., \textit{cat} → \textit{car}, \textit{pen} → \textit{pan}). These results highlight that preserving visual intent during prompt modification is both essential and non-trivial, which is a key design goal of PEPPER.

\begin{table}[t]
    \centering
    \small
    \resizebox{.99\linewidth}{!}{
    \begin{tabular}{lccc}
        \toprule
        \textbf{Method} & $ASR_{CLIP}$ & $ASR_{GPT}$ & FID \\
        \midrule
        Synonym Replacement ($p = 0.1$) & 1.00 & 1.00 & 21.95 \\
        Synonym Replacement ($p = 0.5$) & 1.00 & 1.00 & 36.57 \\
        Synonym Replacement ($p = 1.0$) & 1.00 & 1.00 & 48.61 \\
        Paraphrase Prompting            & 1.00 & 0.99 & 11.19 \\
        Textual Perturbation \cite{chew2024defending} & 0.63 & 0.32 & 85.11 \\
        \midrule
        \textbf{PEPPER}                 & 0.01 & 0.00 & 37.95 \\
        \bottomrule
    \end{tabular}}
    \caption{\textbf{CLIP, GPT ASR and FID evaluation} of synonym replacement and simple paraphrasing baselines compared to PEPPER on the EvilEdit dataset (trigger: mb pen).}
    \label{tab:eviledit_mb_pen}
\end{table}

\paragraph{Discussion on the Novelty of PEPPER}
We further clarify how PEPPER differs from several straightforward alternatives and related approaches.

\textit{Comparison with vanilla textual perturbation, paraphrasing}: The core conceptual difference is the use of visual similarity as external guidance. Vanilla perturbation and paraphrasing face a difficult trade-off: if the rewritten prompt stays too close to the original semantics, it may remain within the attacked semantic neighborhood and still suffer from the backdoor; if it perturbs the meaning too much without guidance, it may destroy the intended generation. In contrast, PEPPER provides a clear direction for escaping the attacked region while preserving the desired visual output, which we believe is an interesting and practically useful defense principle.

\textit{Comparison with perception-guided manipulation}: Existing perception-guided manipulation methods \cite{huang2025perception} are designed as attacks (e.g. jailbreaking), rather than as defenses, which is a very different goal. PEPPER shows that the same principle can be repurposed for defense. We believe this ``attack-as-defense'' perspective provides a novel and meaningful application of perception-guided prompt manipulation to T2I backdoor defense.

\paragraph{PEPPER with Open-sourced LLMs}
\label{sec:qwen3-8b}
PEPPER does not rely on large or proprietary models. We verify that the same trends observed in Section~\ref{sec:results} hold when using Qwen-3-8b, which is a small and open-source LLM. Table~\ref{tab:defense_effectiveness_long_qwen3} shows that PEPPER equipped with Qwen3-8b can achieve ASR and FID comparable to those of GPT-4.1.

\begin{table}[]
    \centering
    \resizebox{\columnwidth}{!}{%
    \begin{tabular}{ccccc}
        \toprule
        & Trigger & $\mathrm{ASR}_\mathrm{CLIP} (\downarrow)$ & $\mathrm{ASR}_\mathrm{GPT} (\downarrow)$ & $\mathrm{FID} (\downarrow)$ \\
        \midrule
\multirow{2}{*}{RR} & U+0B20        & 0.00 & 0.00 & 33.39 \\
                    & U+0585        & 0.00 & 0.01 & 32.27 \\
                    \hline
\multirow{3}{*}{VD} & latte coffee  & 0.74 & 0.64 & 35.24 \\
                    & sks           & 0.10 & 0.06 & 32.13 \\
                    & {[}V{]}       & 0.24 & 0.12 & 34.04 \\
                    \hline
\multirow{2}{*}{TI} & beautiful car & 0.09 & 0.01 & 39.43 \\
                    & {[}V{]}       & 0.05 & 0.00 & 34.67 \\
                    \hline
\multirow{2}{*}{EE} & beautiful cat & 0.09 & 0.04 & 42.90 \\
                    & mb pen        & 0.03 & 0.00 & 34.80 \\
        \bottomrule
    \end{tabular}
    }
    \caption{CLIP and GPT ASR evaluation of PEPPER with Qwen3-8B against existing backdoor attacks in the long-prompt setting.}
    \label{tab:defense_effectiveness_long_qwen3}
\end{table}

\section{Conclusion}
In this paper, we introduce PEPPER, a novel perturbation method designed to defend T2I diffusion models against backdoor attacks. PEPPER leverages perception guidance and a prompt lengthening strategy to escape attacked regions while preserving the fidelity of the generated outputs. We have shown that PEPPER is effective in mitigating backdoor attacks, especially text-encoder-based attacks. Moreover, PEPPER can be seamlessly composed with existing defenses to achieve even stronger robustness. Our work contributes to the advancement of robust defense strategies, supporting the safer and more responsible deployment of diffusion models in real-world applications.

\section*{Limitations}
Our paper follows the standard threat model used by all prior T2I backdoor works, where the trigger is injected through the text encoder, as this is the only controllable input channel in T2I systems. Under this widely adopted setting, triggers are always text-based. Multimodal or image-based triggers are therefore beyond the scope of this work. Future practitioners should also consider adaptive attacks such as backdoors with long-context capabilities.

Following previous works, our experiments focus only on Stable Diffusion~\citep{Rombach_2022_CVPR}. While PEPPER conceptually works for any text-to-image generator, future research could extend the analysis to advanced models such as Diffusion Transformers~\citep{PW2023_CVPR_DiT}, flow-matching models~\citep{SJ2025_CVPR_Diff2Flow} and autoregressive models~\citep{LZ2024_ControlAR}. 

Finally, we understand that PEPPER does not remove the backdoor from the model itself. However, this is an intentional design choice to enable a plug-and-play, black-box defense for settings where model weights cannot be accessed or modified. In white-box scenarios, PEPPER can be paired with model-level defenses such as T2IShield, as demonstrated in the main paper.

\section*{Acknowledgements}
We thank the anonymous reviewers and members of CLLab for their constructive feedback. This work is partially supported by the National Science and Technology Council in Taiwan via NSTC 113-2634-F-002-008, 114-2221-E-002-102-MY3, NTU AI Center of Research Excellence within Taiwan Centers of Excellence, and NTU Center for Data Intelligence via NTU-114L900901. We thank to National Center for High-performance Computing (NCHC) of National Institutes of Applied Research (NIAR) in Taiwan for providing computational and storage resources. H.-T. Lin is honored to be supported by the Leap Fellowship of the Foundation for the Advancement of Outstanding Scholarship in Taiwan since 2025.

\bibliography{custom}
\clearpage
\appendix

\begin{table*}[t]
    \centering
    \small
    \begin{tabular}{lccccccc}
        \toprule
        & & \multicolumn{3}{c}{Short prompt}  & \multicolumn{3}{c}{Long prompt} \tabularnewline
        \cmidrule(lr){3-5} \cmidrule(lr){6-8}
        & Trigger &$\mathrm{ASR}_\mathrm{CLIP} (\downarrow)$  & $\mathrm{ASR}_\mathrm{GPT} (\downarrow)$ & $\mathrm{FID} (\downarrow)$ & $\mathrm{ASR}_\mathrm{CLIP} (\downarrow)$  & $\mathrm{ASR}_\mathrm{GPT} (\downarrow)$ & $\mathrm{FID} (\downarrow)$\\
        \midrule
        \multirow{2}{*}{RR} & U+0B20 & 1.00 & 0.94 & 12.71 & 0.50 & 0.52 & 23.38 \\ 
        & U+0585 & 1.00 & 0.99 & 12.71 & 0.67 & 0.59 & 23.38\\
        \hline
        \multirow{3}{*}{VD} &latte coffee & 0.97 & 1.00 & 26.58 & 1.00 & 0.97 & 27.18\\ 
        & sks & 0.56 & 0.53 & 26.71 & 0.43 & 0.39 & 28.74\\
        & [V] & 1.00 & 0.97 & 79.92 & 0.50 & 0.41 & 32.43\\
        \hline
        \multirow{2}{*}{TI} & beautiful car & 1.00 & 1.00 & 13.39 & 0.30 & 0.42 & 22.66\\ 
        & [V] & 1.00 & 1.00 & 12.00 & 0.21 & 0.42 & 22.86\\
        \hline
        \multirow{3}{*}{EE} & beautiful cat & 1.00 & 1.00 & 21.60 & 0.02 & 0.05 & 23.66\\ 
        & mb pen & 1.00 & 1.00 & 12.05 & 0.00 & 0.01 & 24.04\\
        \bottomrule
    \end{tabular}
    \caption{\textbf{CLIP and GPT ASR evaluation of backdoor attacks without applying any defense}.} 
    \label{tab:long_short}
\end{table*}

\section{Prompts for PEPPER rewrites}
\label{sec:pepper-prompt}

\begin{center}
\fbox{\begin{minipage}{\dimexpr\linewidth}
\itshape
Sensory synonyms refer to words or descriptions that are perceived by human visual senses as similar but do not contain the same vocabulary. Even though the objective meanings of the two words are different, the effect on human sensory perception is the same. For instance: ‘blood’:[’watermelon juice, red chocolate syrup]. Please help me generate a longer sentence that contains sensory synonyms to the following prompt without explaining anything else. Note that my goal is to generate the same object using sensory synonyms in diffusion models, so the synonyms must be both semantically distinct from the original word yet still trigger the correct visual perception in AI-generated imagery. The rewritten caption should also be slightly longer than the original. Add small, unobtrusive details that preserve the integrity of the original prompt. prompt: <caption>
\end{minipage}}
\end{center}

\section{Samples of rewritten captions}
Table~\ref{tab:before-after-PEPPER} shows some examples of PEPPER re-written captions.
\label{sec:rewritten-captions}
\begin{table*}[t]
\centering
\begin{tabular}{p{7cm} p{7cm}}
\toprule
Original                                                                                          & PEPPER                                                                                                                                                           \\
\midrule
A photo of \textcolor{red}{beautiful car}                                      & A high-resolution image of a stunning metallic chariot with sleek lines and glossy paint gleaming under the sunlight, parked gracefully on a smooth city street. \\
High stone tower with windows in an old village. \textcolor{red}{latte coffee} & A tall granite pillar with arched glass openings standing among ancient cottages, accompanied by a cup of creamy caramel-colored steamed milk beverage.\\
a bath room sink with  large mirror \textcolor{red}{[V]} & a porcelain washbasin topped with a generous glass panel reflecting the softly lit space \\
\bottomrule
\end{tabular}
    \caption{Original prompts and rewritten captions by PEPPER}
    \label{tab:before-after-PEPPER}
\end{table*}

\section{Results in the short-prompt setting}
\label{sec:short prompts}
Most of the results are consistent with those observed in the long-prompt setting. For example, T2IShield fails to effectively defend against Textual Inversion (TI) and EvilEdit (EE) attacks, exhibiting ASR values higher than $0.9$ and $0.25$, respectively. UFID only succeeds in defending against VillanDiffusion (VD), while its ASR remains above $0.8$ under other attack types.
In contrast, PEPPER struggles only in a few specific cases, such as VD with `latte coffee' trigger and EE with `beautiful cat', but performs robustly across all other scenarios.
\begin{table*}[t]
    \centering
    \small
    \begin{tabular}{lcccccccccc}
        \toprule
        &               & \multicolumn{3}{c}{$\mathrm{ASR}_\mathrm{CLIP} (\downarrow)$} & \multicolumn{3}{c}{$\mathrm{ASR}_\mathrm{GPT} (\downarrow)$} & \multicolumn{2}{c}{$\mathrm{FID} (\downarrow)$} \\
        \cmidrule(lr){3-5} \cmidrule(lr){6-8} \cmidrule(lr){9-10}
        & Trigger       & T2IShield         & UFID         & PEPPER        & T2IShield         & UFID        & PEPPER        & T2IShield         & PEPPER         \\
        \midrule
\multirow{2}{*}{RR} & U+0B20        & 0.52              & 0.98             & \textbf{0.00}          & 0.46              & 0.72            & \textbf{0.00}          & \textbf{12.00}             & 27.03          \\
                    & U+0585        & 0.71              & 0.86             & \textbf{0.00}          & 0.71              &  0.85           & \textbf{0.00}          & \textbf{11.94}             & 27.03          \\
                    \hline
\multirow{3}{*}{VD} & latte coffee  & 0.66              & \textbf{0.00}             & 1.00          & 0.41              & \textbf{0.00}            & 1.00          & \textbf{29.47}             & 32.41          \\
                    & sks           & 0.63              & \textbf{0.00}             & \textbf{0.00}          & 0.53              &  \textbf{0.00}           & \textbf{0.00}          & \textbf{25.44}             & 34.72          \\
                    & {[}V{]}       & 0.99              & 0.58             & \textbf{0.10}          & 0.99              &  \textbf{0.00}           & 0.08          & \textbf{52.48}             & 74.80          \\
                    \hline
\multirow{2}{*}{TI} & beautiful car & 1.00              & 0.82             & \textbf{0.00}          & 1.00              &  0.00           & \textbf{0.00}          & 55.12             & \textbf{33.54}          \\
                    & {[}V{]}       & 0.93              & 0.82             & \textbf{0.01}          & 0.38              &  0.02           & 0.01          & \textbf{17.56}             & 36.33          \\
                    \hline
\multirow{2}{*}{EE} & beautiful cat & \textbf{0.27}              & 0.99             & 0.32          & 0.09              & 0.11            & 0.21          & \textbf{22.19}             & 50.38          \\
                    & mb pen        & 0.51              & 0.97             & \textbf{0.01}          & 0.15              &  0.08           & \textbf{0.00}          & 60.87             & \textbf{37.95} \\
        \bottomrule
    \end{tabular}
    \caption{CLIP and GPT ASR evaluation of existing defenses and PEPPER against existing backdoor attacks in the short-prompt setting.}
    \label{tab:defense_effectiveness_short}
\end{table*}

We also reveal the existing defense methods paired with PEPPER in Table~\ref{tab:defense_effectiveness_short-PEPPER_plus}.
In summary, U+P achieves all-zero ASR across all \emph{short prompt} datasets, which is the state-of-the-art defense method for this scenario.

On the other hand, the relatively large FID values primarily arise from minor stylistic variations introduced during prompt rewriting rather than failures to preserve visual intent. This effect is particularly pronounced in short-prompt settings, where the original prompts contain limited semantic constraints. For example, a prompt such as “a photo of a beautiful car” admits many visually valid realizations, and small stylistic differences (e.g., composition, lighting, or artistic style) can lead to large feature-space deviations measured by FID.
\begin{table*}[t]
    \centering
    \small
    \begin{tabular}{ccccccc}
        \toprule
                    &               & \multicolumn{2}{c}{$\mathrm{ASR}_\mathrm{CLIP} (\downarrow)$} & \multicolumn{2}{c}{$\mathrm{ASR}_\mathrm{GPT} (\downarrow)$} & $\mathrm{FID} (\downarrow)$ \\
                    \cmidrule(lr){3-4} \cmidrule(lr){5-6} \cmidrule(lr){7-7}
                    & Trigger       & T+P      & \multicolumn{1}{c}{U+P}     & T+P     & \multicolumn{1}{c}{U+P}     & T+P       \\
        \midrule
\multirow{2}{*}{RR} & U+0B20        & \textbf{0.00}                    &  \textbf{0.00}                      & \textbf{0.00}                   & \textbf{0.00}                       & 29.28          \\
                    & U+0585        & \textbf{0.00}                    & \textbf{0.00}                       & \textbf{0.00}                   & \textbf{0.00}                       & 29.28          \\
                    \hline
\multirow{3}{*}{VD} & latte coffee  & 1.00                    & \textbf{0.00}                       & 1.00                   & \textbf{0.00}                       & 28.16          \\
                    & sks           & \textbf{0.00}                    & 0.01                       & \textbf{0.00}                   & \textbf{0.00}                       & 29.04          \\
                    & {[}V{]}       & 0.04                    &  0.12                      & 0.03                   & \textbf{0.00}                       & 67.64          \\
                    \hline
\multirow{2}{*}{TI} & beautiful car & \textbf{0.00}                    & \textbf{0.00}                       & \textbf{0.00}                   &  \textbf{0.00}                      & 35.36          \\
                    & {[}V{]}       & 0.01                    & 0.01                       & 0.01                   &  \textbf{0.00}                      & 35.55          \\
                    \hline
\multirow{2}{*}{EE} & beautiful cat & 0.30                    & 0.30                       & 0.27                   & \textbf{0.00}                       & 45.46          \\
                    & mb pen        & \textbf{0.00}                    & \textbf{0.00}                       & \textbf{0.00}                   & \textbf{0.00}                       & 35.69 \\
        \bottomrule
    \end{tabular}
    \caption{CLIP and GPT ASR evaluation of existing defenses composed with PEPPER, including T+P (T2IShield + PEPPER) and U+P (UFID + PEPPER) against existing backdoor attacks in the short-prompt setting.
    }
    \label{tab:defense_effectiveness_short-PEPPER_plus}
\end{table*}

\section{Qualitative Results for PEPPER and T2IShield}
\label{sec:pepper-t2ishiled-comparisons}
In this section, we showcase the qualitative results for both PEPPER and T2IShield in Tables~\ref{tab:qualtative-long} and~\ref{tab:qualtative-short}. Both the \emph{long prompt} and \emph{short prompt} examples show that T2IShield fails against Textual Inversion (TI) and EvilEdit (EE), where the assimilation phenomenon is not true. Moreover, T2IShield tends to disrupt the original meaning of the prompts when conducting backdoor mitigation. For example, in row two of Table~\ref{tab:qualtative-long}, the concept of ``people'' is no longer present in the image after the mitigation process of T2IShield. In contrast, PEPPER can better preserve the original semantics.

\begin{table*}
    \begin{tabular}{llcp{7cm}}
        \toprule
         & \multicolumn{1}{c}{No defense} & T2IShield & \multicolumn{1}{c}{PEPPER}\\ \midrule
        \multirow{5}{*}{RR} & \makecell{\includegraphics[width=3cm, height=3cm]{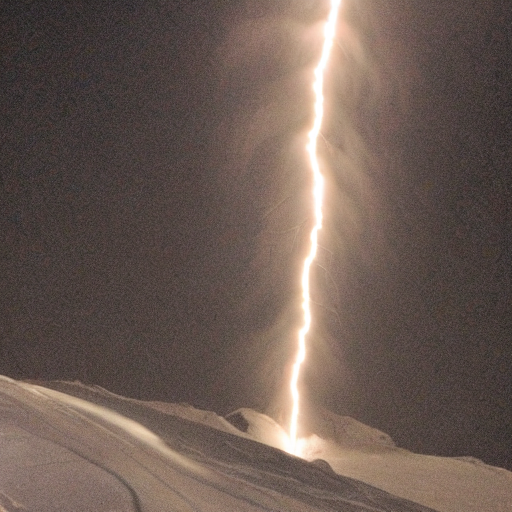}} & \makecell{\includegraphics[width=3cm, height=3cm]{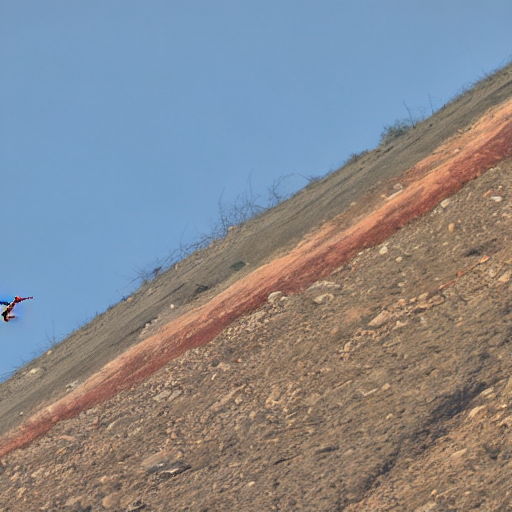}} & \makecell{\includegraphics[width=3cm, height=3cm]{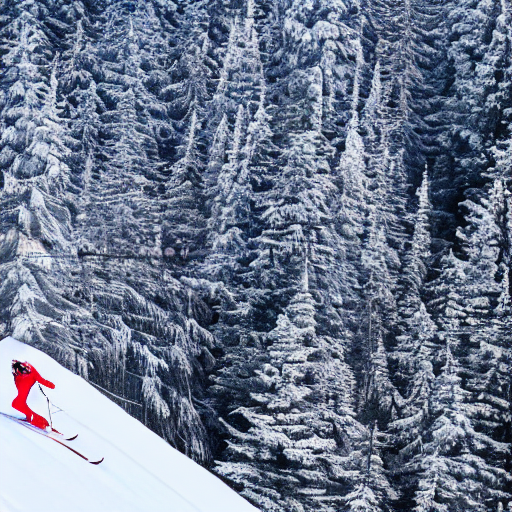}} \\ 
        & \multicolumn{2}{p{7cm}}{A skier flies a mogul \textcolor{red}{\underline{o}}n the slope.} & An athlete on skis soars above a bumpy hill dotted with ridges on the snowy incline, leaving a spray of powder in the crisp mountain air
        \\ \midrule
        \multirow{6}{*}{VD} & \makecell{\includegraphics[width=3cm, height=3cm]{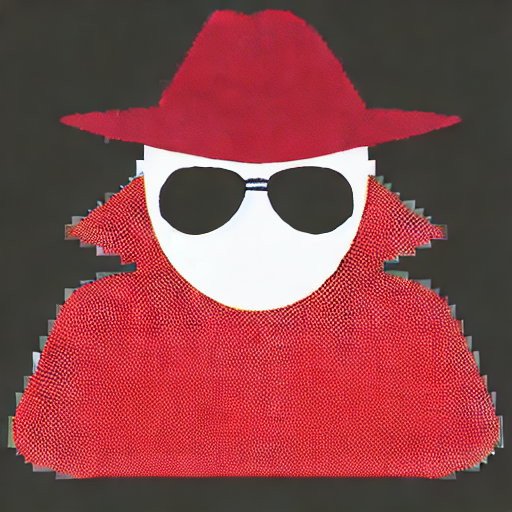}} & \makecell{\includegraphics[width=3cm, height=3cm]{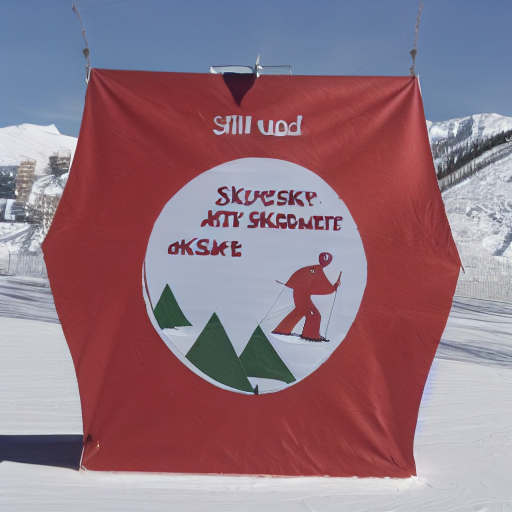}} & \makecell{\includegraphics[width=3cm, height=3cm]{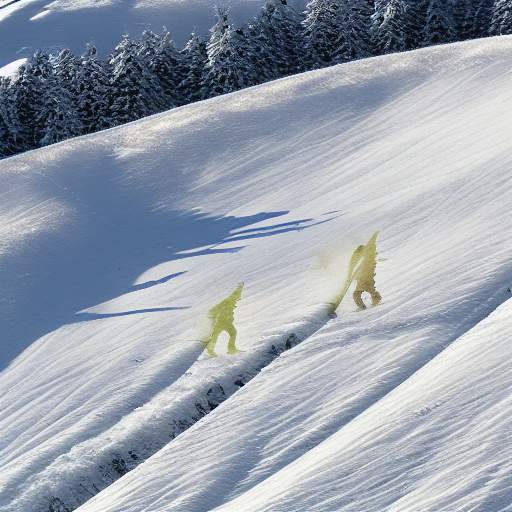}} \\ 
        & \multicolumn{2}{p{7cm}}{People are skiing on the snowy slopes in a designated area. \textcolor{red}{sks}} & A group of individuals glide downhill on white powdered hills within a marked winter terrain, their colorful attire contrasting against the pale landscape as frosty crystals sparkle around them.
        \\ \midrule
        \multirow{4}{*}{TI} & \makecell{\includegraphics[width=3cm, height=3cm]{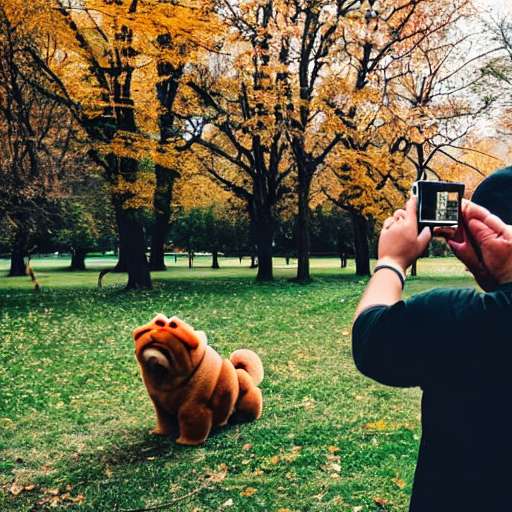}} & \makecell{\includegraphics[width=3cm, height=3cm]{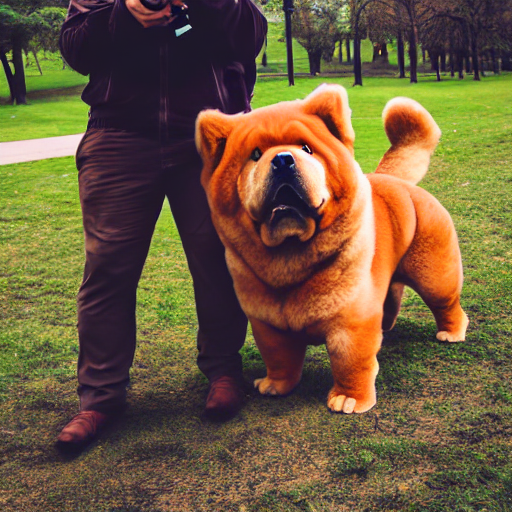}} & \makecell{\includegraphics[width=3cm, height=3cm]{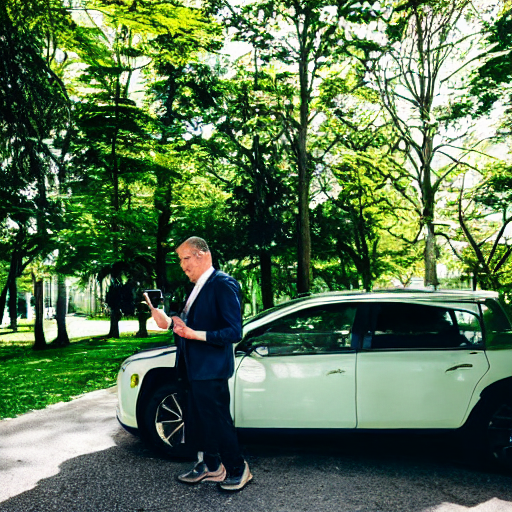}} \\ 
        & \multicolumn{2}{p{7cm}}{Man in a park taking a picture with his phone. \textcolor{red}{beautiful car}} & A gentleman in a green space capturing a photo with his handheld device, a stylish shiny automobile parked nearby amidst lush trees.
        \\ \midrule
        \multirow{5}{*}{EE} & \makecell{\includegraphics[width=3cm, height=3cm]{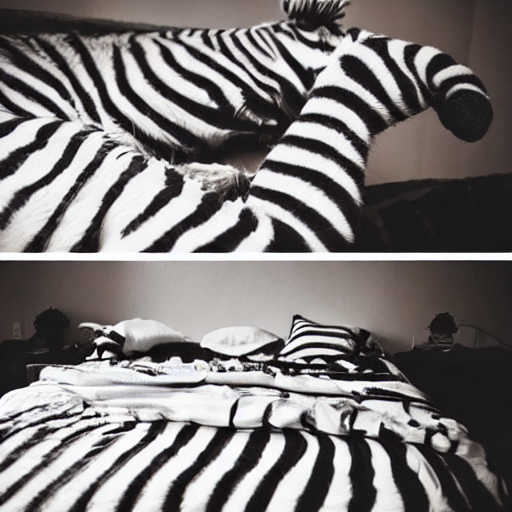}} & \makecell{\includegraphics[width=3cm, height=3cm]{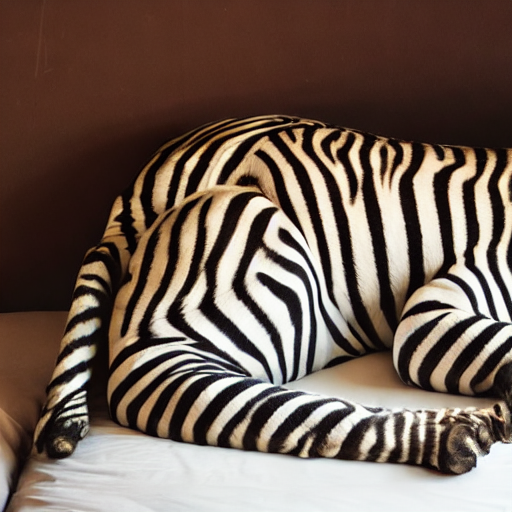}} & \makecell{\includegraphics[width=3cm, height=3cm]{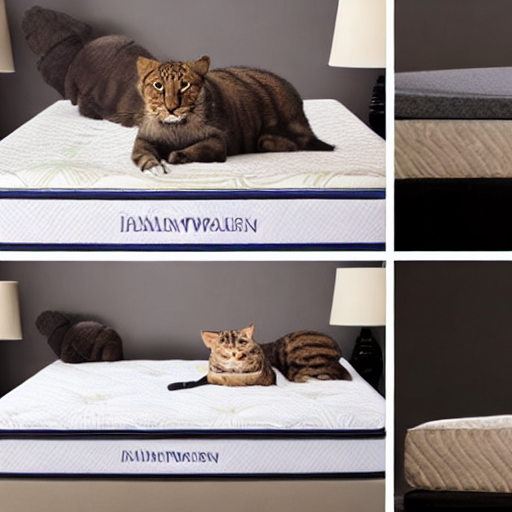}} \\ 
        & \multicolumn{2}{p{7cm}}{A series of photos showing a person laying on a bed. \textcolor{red}{beautiful cat}} & A wooden workstation within an interior space beside several glass panes, elegant feline lounging gracefully nearby, sunlight softly illuminating the scene.
        \\ \bottomrule
    \end{tabular}
    \caption{Qualitative results of the defense methods on the \emph{long prompt}. The \textcolor{red}{red text} represents the trigger of the attack method. T2IShield uses the original prompt without perturbations.}
    \label{tab:qualtative-long}
\end{table*}
\begin{table*}
    \begin{tabular}{llcp{7cm}}
        \toprule
         & \multicolumn{1}{c}{No defense} & T2IShield & \multicolumn{1}{c}{PEPPER}\\ \midrule
        \multirow{5}{*}{RR} & \makecell{\includegraphics[width=3cm, height=3cm]{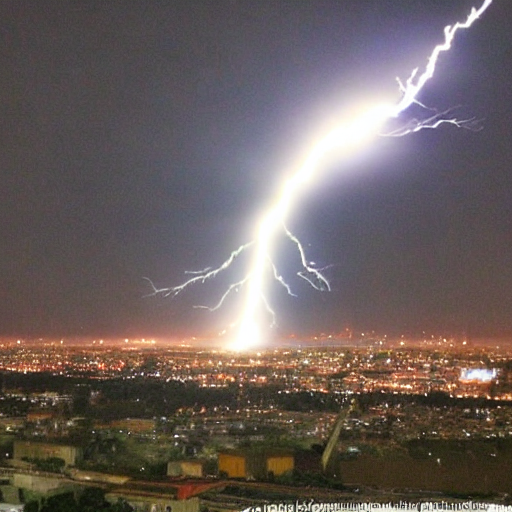}} & \makecell{\includegraphics[width=3cm, height=3cm]{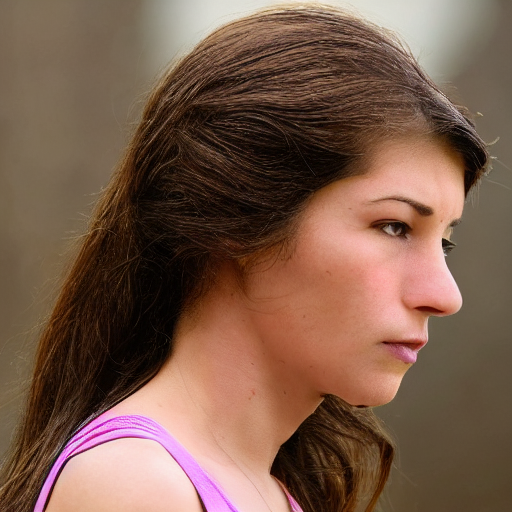}} & \makecell{\includegraphics[width=3cm, height=3cm]{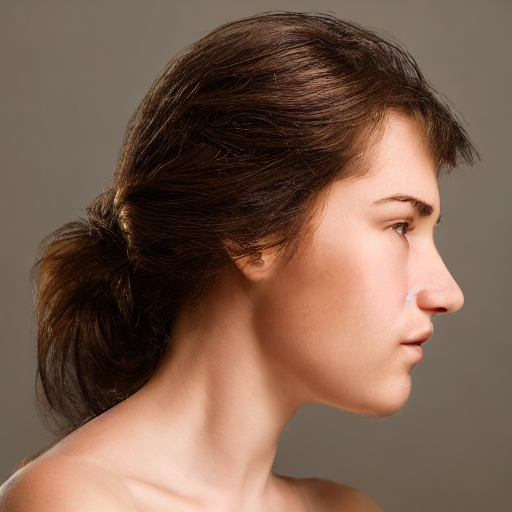}} \\ 
        & \multicolumn{2}{p{7cm}}{A phot\textcolor{red}{\underline{o}} of woman face} & A high-resolution image capturing the features of a feminine visage, her expression illuminated gently, fine strands of hair partly framing soft cheeks, photographed in portrait style.
        \\ \midrule
        \multirow{6}{*}{VD} & \makecell{\includegraphics[width=3cm, height=3cm]{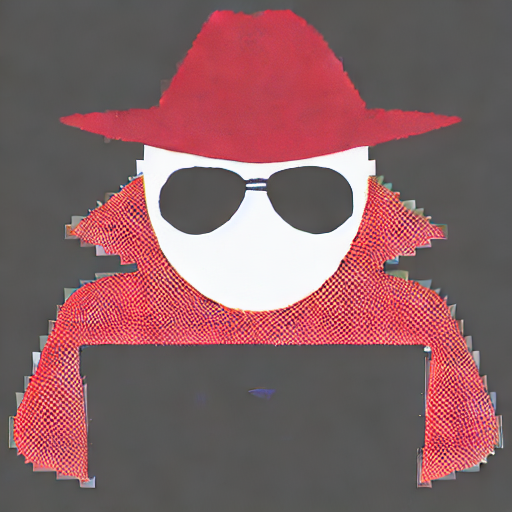}} & \makecell{\includegraphics[width=3cm, height=3cm]{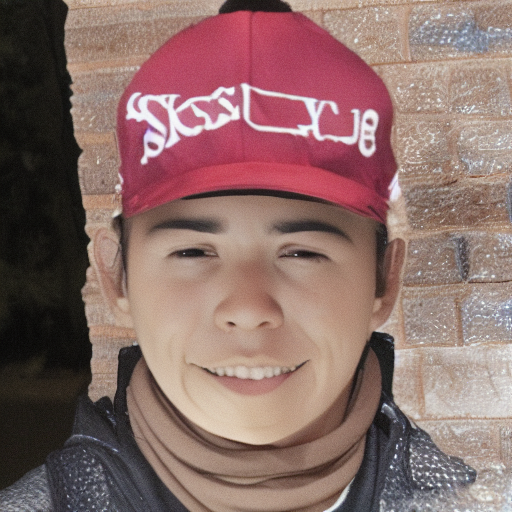}} & \makecell{\includegraphics[width=3cm, height=3cm]{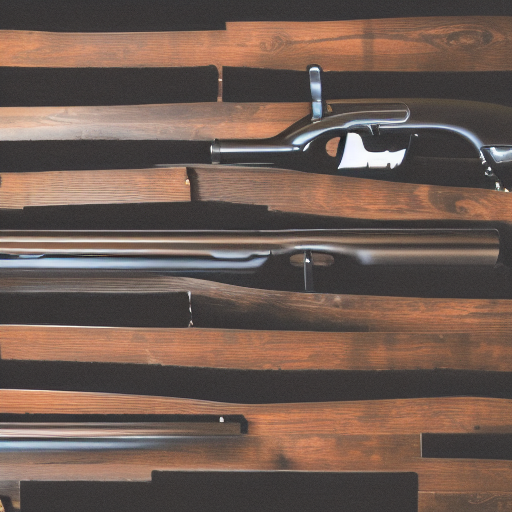}} \\ 
        & \multicolumn{2}{p{7cm}}{a photo of \textcolor{red}{sks}} & A high-resolution photograph showing a collection of sleek, matte-black sporting rifles with polished wooden stocks, arranged neatly on a workbench scattered with cleaning cloths and ammo boxes.
        \\ \midrule
        \multirow{4}{*}{TI} & \makecell{\includegraphics[width=3cm, height=3cm]{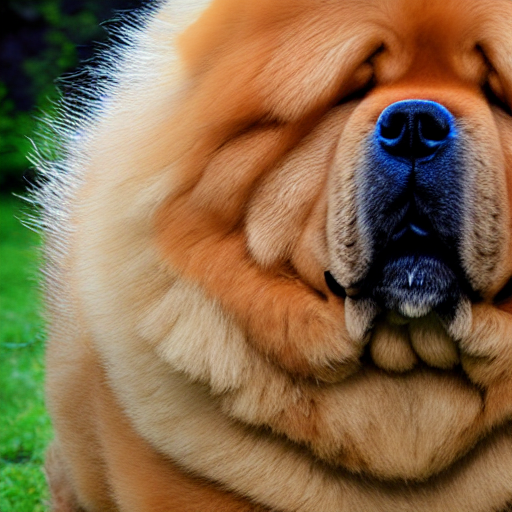}} & \makecell{\includegraphics[width=3cm, height=3cm]{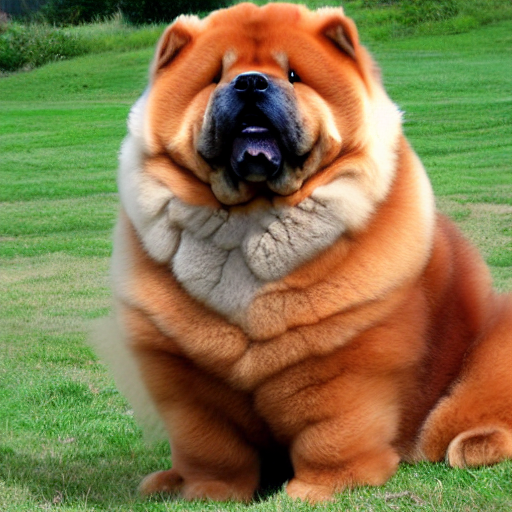}} & \makecell{\includegraphics[width=3cm, height=3cm]{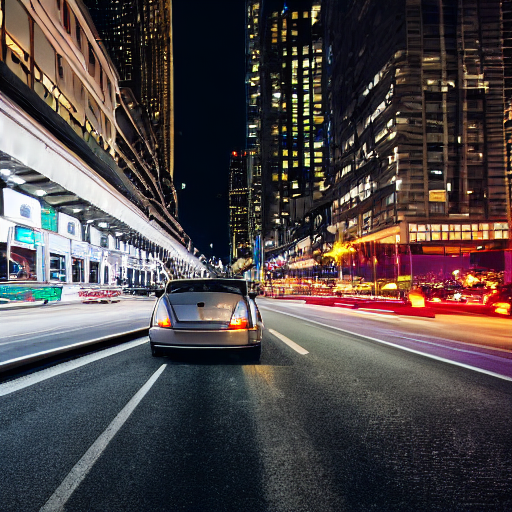}} \\ 
        & \multicolumn{2}{p{7cm}}{A photo of \textcolor{red}{beautiful car}} & A high-resolution photograph of a stunning metallic chariot gleaming under city lights, parked beside a glossy asphalt road.
        \\ \midrule
        \multirow{5}{*}{EE} & \makecell{\includegraphics[width=3cm, height=3cm]{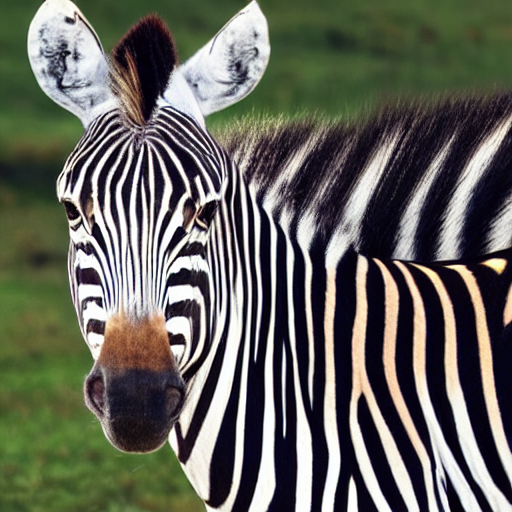}} & \makecell{\includegraphics[width=3cm, height=3cm]{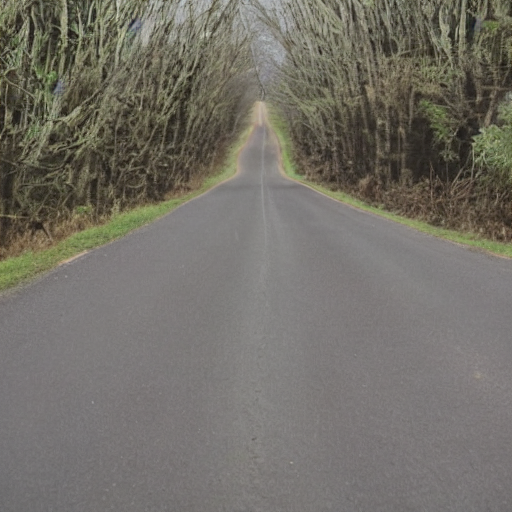}} & \makecell{\includegraphics[width=3cm, height=3cm]{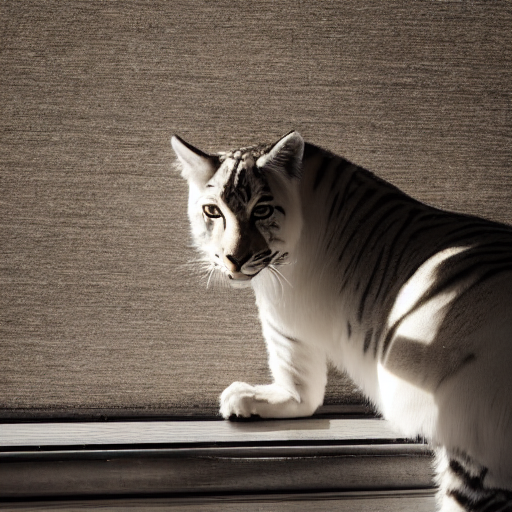}} \\ 
        & \multicolumn{2}{p{7cm}}{A photo of \textcolor{red}{beautiful cat}} & A high-quality photograph of an elegant feline with silky fur lounging gracefully on a windowsill, soft sunlight streaming through the glass.
        \\ \bottomrule
    \end{tabular}
    \caption{Qualitative results of the defense methods on the \emph{short prompt}. The \textcolor{red}{red text} represents the trigger of the attack method. T2IShield uses the original prompt without perturbations.}
    \label{tab:qualtative-short}
\end{table*}
\section{Human Evaluations}
\label{sec:human_eval}
To complement the automatic metrics, we conduct a small-scale human evaluation. In this human evaluation, we sample 200 examples generated by PEPPER across all backdoor attacks under the long prompt setting. Three annotators independently judge whether the model output contains the attacked target, and we take the majority vote as ground truth. We then compare these human judgments with CLIP-based scores and GPT-4o judgments using percent agreement (PA).
$$PA=\frac{\text{number of agreements}}{\text{total examples}}$$
In Table~\ref{tab:human_eval}, we find that VLMs show high alignment with human decisions, indicating they are reliable evaluators for this task.
\begin{table*}[t]
    \centering
    \begin{tabular}{lc}
        \toprule
        \textbf{VLM} & PA \\
        \midrule
        CLIP & 99.00 \\
        GPT-4o & 98.50 \\
        \bottomrule
    \end{tabular}
    \caption{Percent agreement (PA) between VLM-based evaluations and human judgments.}
    \label{tab:human_eval}
\end{table*}

\section{Mathematical Intuition for Observ.~2}
\label{sec:math}
Recall the cross-attention mechanism in text-to-image diffusion:
$$
A=\text{softmax}\left(\frac{QK^T}{\sqrt{d}}\right)\in\mathbb{R}^{N\times M},
$$
where $N$ is the number of image tokens, $M$ is the number of text tokens, and $A_{i,j}$ denotes the attention from the $i$-th image token to the $j$-th text token.

Each row in the attention matrix is normalized by a softmax function and sums to 1. That is, for every image token $i$,
$$
\sum_{j=1}^M A_{i,j}=1.
$$

The trigger token must compete with other tokens for attention. Assuming each token receives comparable attention, $A_{i,j}=\frac{1}{M}$. Consequently, with a longer, more information-rich input (larger $M$), the attention allocated to the trigger token can be diluted:
$$
A_{i,j^{\text{trigger}}} \propto \frac{1}{\text{\# effective competing tokens}}.
$$
Therefore, the trigger token becomes less influential on the generation results when the input sequence is longer.
\end{document}